\documentclass[acmtog]{acmart}

\usepackage{booktabs} % For formal tables

\usepackage{pifont}
\usepackage{comment}
\usepackage{multirow}
\usepackage{color,colortbl}
\usepackage{makecell}
\usepackage{stfloats}

\newcommand{\OURS}{DiffCAD}

\newcommand{\cmark}{\ding{51}}%
\newcommand{\xmark}{\ding{55}}%

\def\ScaleFeatureExtractor{\mathcal{E}_s}
\def\DepthFeatureExtractor{\mathcal{E}_n}
\def\NOCSFeatureEmbedding{\mathcal{P}_z}

\def\CADEncodingModel{\Phi_o}
\def\SceneScaleDiffusionModel{\Phi_s}
\def\CADAlignmentDiffusionModel{\Phi_n}
\def\CADRetrievalDiffusionModel{\Phi_z}
\definecolor{tabbestcolor}{rgb}{0.785, 0.851, 0.969}

\usepackage[ruled]{algorithm2e} % For algorithms

\SetAlFnt{\small}
\SetAlCapFnt{\small}
\SetAlCapNameFnt{\small}
\SetAlCapHSkip{0pt}

\setcopyright{acmlicensed}
\acmJournal{TOG}
\acmYear{2024} 
\acmVolume{43} 
\acmNumber{4} 
\acmArticle{106} 
\acmMonth{7}
\acmDOI{10.1145/3658236}

\begin{document}

\title{\OURS: Weakly-Supervised Probabilistic CAD Model Retrieval and Alignment from an RGB Image}

\author{Daoyi Gao}
\orcid{0000-0003-0458-8107}
\affiliation{
 \institution{Technical University of Munich}
 \country{Germany}
 }
\email{daoyi.gao@tum.de}

\author{D\'avid Rozenberszki}
\affiliation{
 \institution{Technical University of Munich}
 \country{Germany}
 }
\email{david.rozenberszki@tum.de}

\author{Stefan Leutenegger}
\affiliation{
 \institution{Technical University of Munich}
 \country{Germany}
 }
\email{stefan.leutenegger@tum.de}

\author{Angela Dai}
\affiliation{
 \institution{Technical University of Munich}
 \country{Germany}
 }
\email{angela.dai@tum.de}

\renewcommand\shortauthors{Gao, D. et al}

\begin{teaserfigure}
  \includegraphics[width=\textwidth]{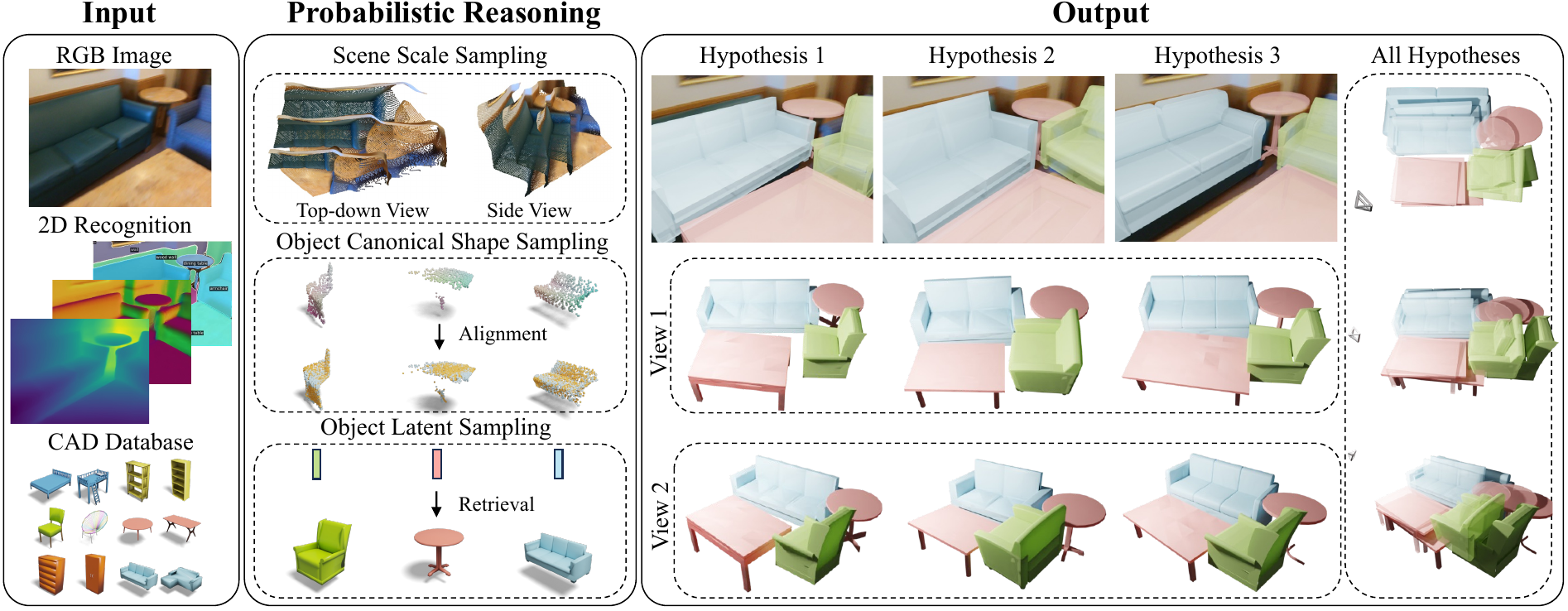}
  \caption{
  We introduce \OURS, a weakly-supervised probabilistic approach for CAD model retrieval and alignment to a single RGB image.
  Unlike existing methods that require expensive CAD associations to real images for supervision, our approach is only trained on synthetic data, yet it can demonstrate superior performance on real-world datasets. 
  }
  \label{fig:teaser}
\end{teaserfigure}

\begin{abstract}

Perceiving 3D structures from RGB images based on CAD model primitives can enable an effective, efficient 3D object-based representation of scenes.
However, current approaches rely on supervision from expensive yet imperfect annotations of CAD models associated with real images, and encounter challenges due to the inherent ambiguities in the task -- both in depth-scale ambiguity in monocular perception, as well as inexact matches of CAD database models to real observations.
We thus propose \OURS, the first weakly-supervised probabilistic approach to CAD retrieval and alignment from an RGB image.
We learn a probabilistic model through diffusion, modeling likely distributions of shape, pose, and scale of CAD objects in an image.
This enables multi-hypothesis generation of different plausible CAD reconstructions, requiring only a few hypotheses to characterize ambiguities in depth/scale and inexact shape matches.
Our approach is trained only on synthetic data, leveraging monocular depth and mask estimates to enable robust zero-shot adaptation to various real target domains.
Despite being trained solely on synthetic data, our multi-hypothesis approach can even surpass the supervised state-of-the-art on the Scan2CAD dataset by $5.9\%$ with 8 hypotheses.

\end{abstract}




\maketitle

\section{Introduction}
\label{sec:intro}

Obtaining 3D perception from 2D input is a fundamental challenge within the realm of computer graphics, carrying extensive implications for various applications in virtual environments and digital content creation, such as VR applications and the development of interactive gaming experiences.

While 2D perception from a single RGB image has achieved significant success in recent years \cite{russakovsky2015imagenet, he2017mask, ranftl2021vision, kirillov2023segment}, 3D perception -- required for enabling tasks such as exploration and interaction with objects -- remains challenging, requiring not only object recognition but reconstruction of diverse, complex objects. 
In 3D perception, leveraging a CAD model basis for representing 3D objects in an image enables the use of a strong 3D prior on geometric structures and generates clean, compact mesh outputs directly compatible with efficient modern rendering pipelines, thereby enhancing compatibility for subsequent graphics tasks.

While such CAD retrieval and alignment to generate an object-based 3D scene representation of real-world scene observation has shown strong promise~\cite{kuo2020mask2cad,kuo2021patch2cad,izadinia2017im2cad,avetisyan2019scan2cad,maninis2022vid2cad, gumeli2022roca, langer2022sparc}, existing methods tend to rely on expensive real-world annotations that necessitate trained annotators~\cite{avetisyan2019scan2cad}. Moreover, these annotations fail to provide exact ground truth matches regarding shape and pose, as existing CAD databases cannot tractably cover the distribution of general real objects. As shown in the bottom row of Fig.~\ref{figure:qualitative_mainpaper}, the chair's footrest is incorrectly annotated as a table. In contrast, the synthetic domain offers precise correspondences between renderings and object shapes inherent in the data without requiring manual annotations. Thus, rather than relying on limited, imperfect real-world data annotations for supervision, we adopt a synthetic-to-real domain adaptation strategy~\cite{ze2022category,schwonberg2023survey} to propose a weakly-supervised approach trained only on synthetic data, that can be applied to various real-world images.

To effectively address the problem of CAD retrieval and alignment with weak supervision, our key insight is that a probabilistic model is required. This enables capture of not only ambiguities in synthetic to real domain transfer, but also ambiguities inherent in the monocular perception task, such as depth-scale ambiguity and object shape ambiguity arising from partial visibility. Notably, previous methods overlook the inherent probabilistic nature of these challenges, opting for deterministic inference. In addressing the inherent ambiguities in single-view 3D perception systems, we model the likely distributions of scene scale, object pose, and object shape as separate and disentangled conditional generative tasks. To this end, we introduce \OURS{}, the first probabilistic approach for CAD retrieval and alignment to an RGB image without real-world supervision.

Our approach first employs a diffusion model to analyze potential scene scales based on an estimated monocular depth map. 
This enables considering multiple feasible solutions for object translation and scaling when subsequently solving for object poses. 
Given the estimated scales and depth map, we then model object pose through diffusion to predict an explicit canonical representation of objects, which we parameterize as normalized object coordinates (NOCs)~\cite{wang2019normalized}. The NOCs inform robust pose estimation and guide the diffusion of an object shape descriptor for CAD retrieval.

At inference time, we consider RGB images from real scenes. 
To help reduce the domain gap, we operate on machine-generated monocular depth and mask estimates from the RGB image (as photorealistic RGB generation from synthetic 3D data is challenging and expensive).
We can then sample from our learned distributions for multiple plausible CAD reconstruction results.
Our sampling scheme cascades from the scene-level diffusion, which offers potential scene scales. It then progresses to the explicit object representation diffusion, generating multiple NOCs that align with the scaled depth maps. Finally, it ends with the object latent diffusion, where the sampled latent vectors are used to query the CAD model database. 
This sampling scheme enables the generation of several CAD objects and poses that capture shape and depth-scale ambiguity in representing the 3D scene depicted in an image.

Experiments on ScanNet~\cite{dai2017scannet, avetisyan2019scan2cad} images show that our learned probabilistic distributions well-capture likely CAD-based reconstructions, and with only 8 hypotheses, can even outperform fully-supervised state-of-the-art by $5.9\%$.

In summary, our contributions are: 

\begin{itemize}
    \item We propose the first probabilistic approach to CAD model retrieval and alignment from an RGB image, capturing inherent ambiguities due to depth-scale and lack of exact CAD matches in a database.
    \item We formulate our learned probabilistic model with diffusion processes that capture the distribution of scene scale, object pose, and object shape, with efficient sampling for multiple plausible hypotheses of CAD reconstructions.
    \item Our probabilistic approach leverages machine-estimated depth and 2D masks, enabling robust generalization to real images while training only on synthetic data. 
\end{itemize}

\begin{figure*}[!t]
      \centering
      \includegraphics[width=0.95\textwidth]{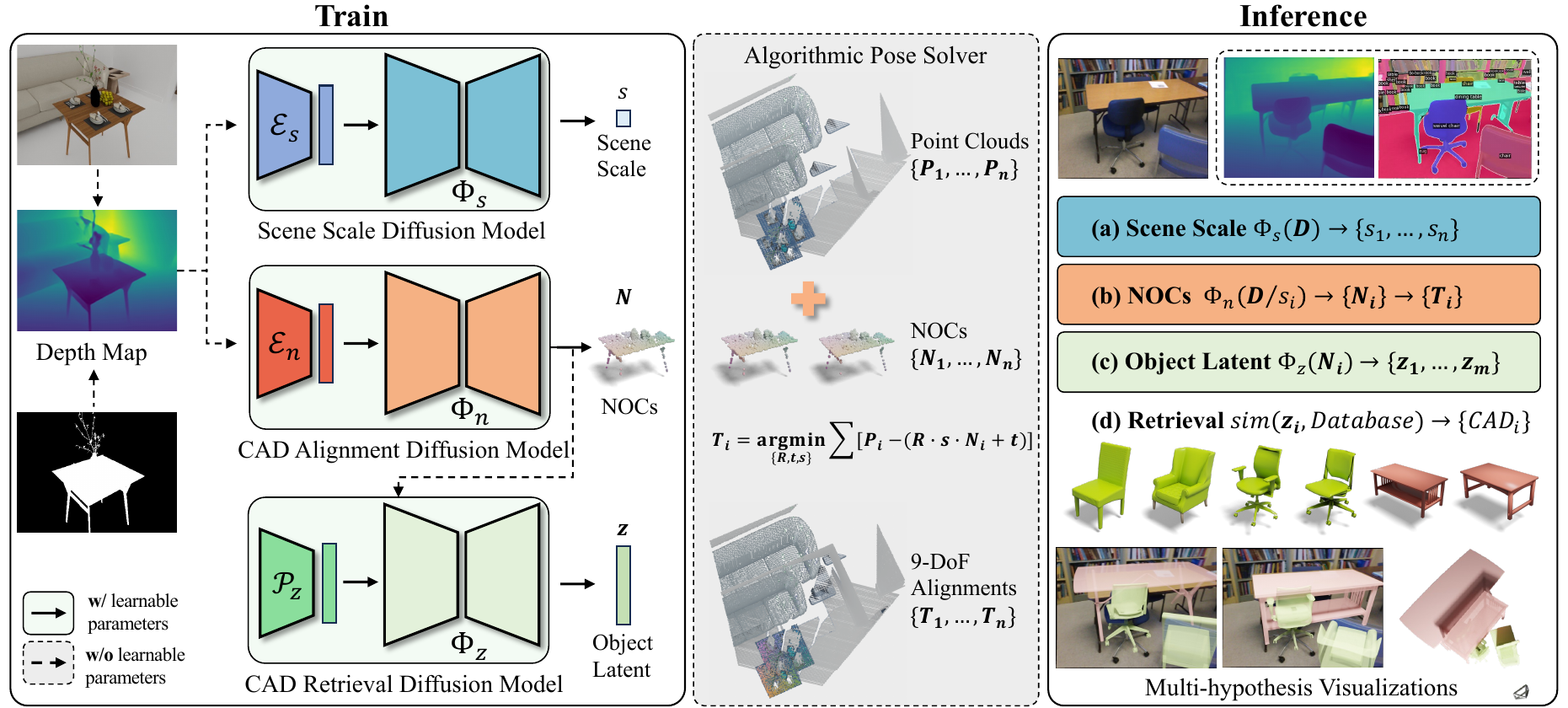}
      \caption{
      \textbf{Method Overview}. To facilitate multi-hypothesis reasoning for CAD model retrieval and alignment to a single image, we employ diffusion modeling over scene scale, object pose, and shape.
      From an input RGB image, we employ machine-generated estimates of depth and instance segmentation.
      From the estimated depth, we estimate scene scales with  $\SceneScaleDiffusionModel$. $\CADAlignmentDiffusionModel$ uses the back-projected estimated depth of each detected object to output hypotheses for its Normalized Object Coordinates (NOCs). $\CADRetrievalDiffusionModel$ then uses the estimated NOCs to predict the object shape as a latent vector that can be used for retrieval.
      Our probabilistic modeling also enables robust real-world CAD retrieval and alignment while training only on synthetic data.
      }
      \label{figure:pipeline}
\end{figure*}

\section{Related Work}
\label{sec:related}

\subsection{2D Object Perception} 
With recent advancements in deep learning, significant progress has been achieved in various aspects of 2D object perception, including object detection~\cite{lin2017feature, he2017mask, Carion2020EndtoEndOD, ge2021yolox, Fang2022EVAET}, instance segmentation~\cite{he2017mask, ghiasi2021open, Li2022LanguagedrivenSS, Liang2022OpenVocabularySS, xu2023open,kirillov2023segment}, and reasoning about object and scene geometry, such as metric depth estimation~\cite{Bae2022IronDepthIR, Ramamonjisoa2019SharpNetFA, Li2022DepthFormerEL, Bhat2020AdaBinsDE, bhat2023zoedepth, tri-zerodepth}, and object normal map estimation~\cite{eftekhar2021omnidata, Bae2022IronDepthIR}. Our approach builds upon this progress in 2D recognition, leveraging segmentation and depth estimates to build a 3D understanding of the scene through probabilistic reasoning regarding scene scale, object pose, and shape.

\subsection{Single-View Object Reconstruction}
The task of inferring object shapes from images through reconstruction has gained significant attention in recent years. Researchers have explored a variety of explicit and implicit object representations, including voxel grids~\cite{choy20163d, Wan20193DMaskGANUnsupervisedS3}, point clouds~\cite{fan2017point, Mandikal20183DLMNetLE, Zeng2022LIONLP}, polygonal meshes~\cite{Wang2018Pixel2MeshG3, gkioxari2019mesh, Pan2019DeepMR}, and neural fields~\cite{mescheder2019occupancy, Yu2020pixelNeRFNR, Lin2022VisionTF, Deng2022NeRDiSN}. 
Recent methods have also leveraged such single-view object reconstruction to construct scenes on an object basis~\cite{Nie2020Total3DUnderstandingJL, Irshad2022CenterSnapSM, Zhang2021Holistic3S, Liu2022TowardsHS}.  
These approaches typically train on synthetic 3D shape data, producing flexible, underconstrained output representations, resulting in reconstructions that are typically over-tessellated (due to Marching Cubes) and often exhibit local noise (e.g., imperfect flat surfaces) or missing finer-grained structures, due to the high dimensionality of the output representation. In contrast, our approach leverages a stronger 3D prior by employing CAD model databases to directly model scene geometry, resulting in plausible object reconstructions up to the fidelity of the CAD database, and a compact representation directly suitable for downstream applications.

\subsection{CAD Model Retrieval and Alignment}
As CAD representations enable efficient, mesh-based representations of a scene, various methods have explored CAD retrieval and alignment to real-world RGB images \cite{izadinia2017im2cad,kuo2020mask2cad,kuo2021patch2cad,gumeli2022roca,langer2022sparc} as well as RGB video~\cite{maninis2022vid2cad} and  RGB-D scans \cite{avetisyan2019scan2cad,avetisyan2019end,avetisyan2020scenecad, Di2023UREDU3, Beyer2022WeaklySupervisedEC, nan2012search, kim2012acquiring, shao2012interactive, li2015database}. 

For the challenging task of single-view RGB input, early methods have relied on the availability of ground truth scale information~\cite{kuo2020mask2cad, kuo2021patch2cad} for full 9-DoF 3D reconstruction. Recent works have also proposed directly learning metric depth estimation within the target domain~\cite{gumeli2022roca, langer2022sparc, langer2023sparse}.
These methods all not only rely on ground truth supervision in the target domain, which is expensive and inexact to acquire, but produce deterministic outputs, whereas the task setting is inherently ambiguous due to depth-scale ambiguity and inexact CAD matches. 
In contrast, we propose a probabilistic approach to effectively model multiple plausible hypotheses, and maintain general applicability to real images while trained only on synthetic data.

\subsection{Diffusion Models}
Recently, diffusion models \cite{SohlDickstein2015DeepUL, Ho2020DenoisingDP, Song2020ScoreBasedGM, Kingma2021VariationalDM, rombach2021highresolution} have shown remarkable success in modeling image generation while employing an implicit probabilistic model. 
Due to their success in generative modeling of RGB images, recent works have shown more general applicability to various tasks,  including segmentation~\cite{Baranchuk2021LabelEfficientSS, xu2023open}, keypoint matching~\cite{tang2023dift, luo2023dhf, hedlin2023unsupervised}, retrieval-based image generation~\cite{blattmann2022retrieval, sheynin2022knn, Chen2022ReImagenRT}, and 3D shape generation~\cite{Zhou20213DSG, Zeng2022LIONLP, Nam20223DLDMNI, Zhang20233DShape2VecSetA3, Li2022DiffusionSDFTV, Chou2022DiffusionSDFCG, shue20233d, Koo2023SALADPL, erkocc2023hyperdiffusion}.
We propose to employ diffusion modeling to characterize probabilistic models of scenes, along with CAD shape and pose for CAD model reconstruction of a single-view image.

\section{Method}
\label{sec:method}

\subsection{Overview}
\label{sec:method_overview}
Given an RGB image $\mathbf{I}$ and a database of $n$ CAD models $\{O_i\}_{i\in [1,n]}$, our goal is to represent objects in the scene by combining CAD models from the database with a set of 9-DoF transformations $\{\mathbf{T}_i\}_{i\in [1,m]}$ that align these models to the metric camera space, forming a compact, object-based 3D scene representation.

Figure~\ref{figure:pipeline} shows an overview of our method. We propose a combination of disentangled diffusion models to probabilistically model ambiguities in monocular perception and shape matching. Specifically, we model scene scale $\SceneScaleDiffusionModel$, pose prediction $\CADAlignmentDiffusionModel$, and CAD model retrieval $\CADRetrievalDiffusionModel$ as diffusion processes to effectively capture likely CAD model reconstructions of an image.

To reason robustly across various domains, we operate on machine-generated depth $\mathbf{D}$ and semantic instance masks estimated from $\mathbf{I}$, denoted as $\mathbf{D}_p$ indicating the masked depth estimate of each object (object index left out for simplicity of notation). 

We start by addressing the scene-level depth-scale ambiguity. We learn the distribution of potential scales within the scene with a diffusion model $\SceneScaleDiffusionModel$, conditioned on $\mathbf{D}$. 
We can then sample scales $\{s_i\}_{i\in [1,n]}$ from $\SceneScaleDiffusionModel$, to produce the most likely metric scenes $\{\mathbf{D}_i= \frac{1}{s_i}\mathbf{D}\}_{i\in [1,n]}$. 
This enables us to facilitate multiple reasonable transformations of objects.

\begin{figure}[hb]
      \centering
      \includegraphics[width=0.95\columnwidth]{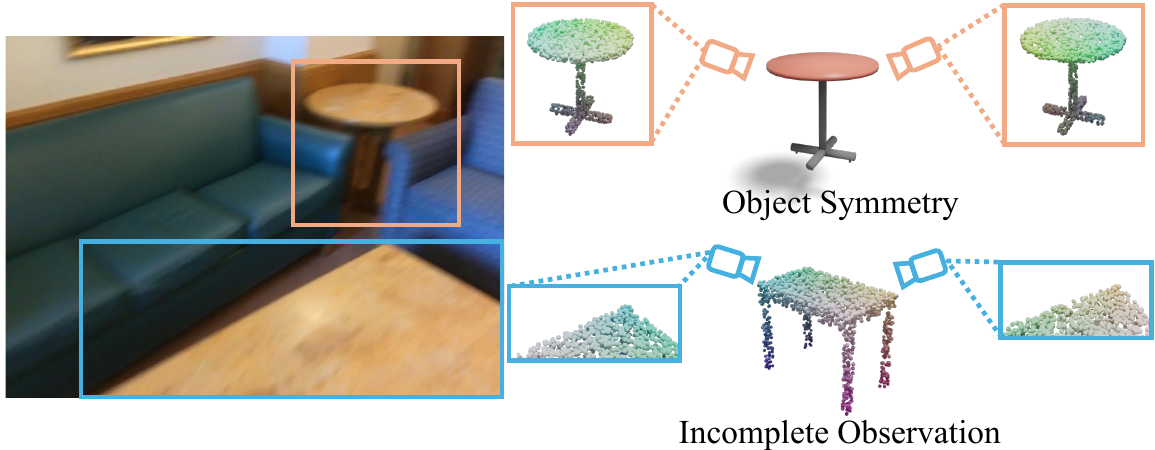}
      \caption{
      \textbf{Multi-hypothesis nature of NOC.} The symmetry in object geometry and the incomplete perception can lead to multiple feasible alignments, which we characterize in our probabilistic, diffusion-based approach.
      }
      \label{figure:multinocs}
\end{figure}

We then model the object pose with diffusion model $\CADAlignmentDiffusionModel$, which estimates the normalized object coordinates (NOCs)~\cite{wang2019normalized} conditioned on the back-projected point clouds $\mathbf{P}$ of $\mathbf{D}_p$. This probabilistic formulation accounts for known multi-hypothesis issues for NOCs~\cite{zhang2024generative}, as a single observation can lead to multiple possible solutions due to the object's symmetric structure or incomplete 2D observations, as shown in Fig.~\ref{figure:multinocs}. The 9-DoF pose is recovered through RANSAC~\cite{fischler1981random} given NOC and $\mathbf{P}$. 

Given the estimated object pose characterized through NOCs, we also predict the object shape represented as a latent shape code through diffusion model $\CADRetrievalDiffusionModel$.
This probabilistic model enables capturing a distribution of possible matches, which enables more robust shape retrieval in real scenarios where no exact match from the database exists.

\subsection{Diffusion Models}
This work follows the denoising diffusion probabilistic models (DDPMs) formulation~\cite{Ho2020DenoisingDP} to model CAD model retrieval and alignment from a single image, including the scene scale $s$, NOCs $\mathbf{N}$, and the latent object shape $\mathbf{z}$.

Given a data distribution $q(x)$, we denote a sampled data point as $x_0$. The \textit{forward process} iteratively injects Gaussian noise to transform $x_0$ to $x_T$, which fits an isotropic Gaussian in $T$ timesteps, by a Markovian process~\cite{Ho2020DenoisingDP, song2019generative}:
\begin{align} 
    q(x_t|x_{t-1}) &= \mathcal{N}(x_t; \sqrt{1 - \beta_t} x_{t-1}, \beta_t \mathbf{I}), \label{eq:q1}\\
    q(x_{1:T}|x_0) &= \prod_{t=1}^{T} q(x_t|x_{t-1}), \label{eq:q2}
\end{align}
where $t \in \left[1,T\right]$ and $\beta_t$ is a pre-defined variance schedule.

The denoising neural network $\Phi$ learns to fit $p_{\Phi}(x_{t-1}|x_t)$ that can approximate $q(x_t|x_{t-1}, x_0)$ to and thus recover the $x_0$ by the \textit{reverse process}:
\begin{align} 
p_{\Phi}(x_{t-1}|x_t) &= \mathcal{N}(x_{t-1}; \mu_{\Phi}(x_t, t), \Sigma_{\Phi}(x_t, t)),  \label{eq:p1}\\
p_{\Phi}(x_{0:T}) &= p(x_T) \prod_{t=1}^{T} p_{\Phi}(x_{t-1}|x_t).  \label{eq:p2}
\end{align}

We can thus train our diffusion models $\SceneScaleDiffusionModel$, $\CADAlignmentDiffusionModel$, and $\CADRetrievalDiffusionModel$ following $\epsilon$-formulation~\cite{,Ho2020DenoisingDP, rombach2021highresolution} or directly optimize for $x_0$ following~\cite{ramesh2022hierarchical, chou2023diffusion} to recover individual data distributions.

\subsection{Scene Scale Diffusion}
\label{sec:scale_diffusion}
As monocular perception inherently contains depth-scale ambiguity, we use $\SceneScaleDiffusionModel$ to model likely scene scales given machine-generated depth $\mathbf{D}$. 
% Since we condition on a machine-generated depth $\mathbf{D}$, we model this as the scale difference from the predicted depth.
Given the inherent bias in monocular depth estimates for both synthetic and real data, we argue that modeling the distribution of the scale factor between the prediction and the ground truth is more robust across domains. 
We define the target scale difference between the predicted depth map $\mathbf{D}$ and the reference depth map $\mathbf{D}_\textrm{gt}$ as follows:
\begin{equation}
s_{gt} = \text{avg} \Big( \frac{\mathbf{D} \odot \mathbf{M}}{\mathbf{D}_\textrm{gt} \odot \mathbf{M}} \Big).
\label{eq:scale_def}
\end{equation}
Here, $\mathbf{M}$ represents the mask of the target object, and $\odot$ denotes the Hadamard product.

$\SceneScaleDiffusionModel$ is conditioned on features extracted from estimated depth map $\mathbf{D}$ from a pre-trained ResNet-50-FCN backbone~\cite{long2015fully}.
Since the defined scene scale target is an isotropic scalar that zooms the initial estimates uniformly in each direction, we construct the scale diffusion target vector $\mathbf{S} $ with each element equals to $s_\textrm{gt}$ and has the same size as the depth feature map, and concatenate them as input to the model. We adopt the objective function proposed in~\cite{Ho2020DenoisingDP} for training the diffusion U-Net $\SceneScaleDiffusionModel$:

\begin{equation}
\mathcal{L}_{s} = \mathbb{E}_{\epsilon \sim \mathcal{N}(0,I), t} \Big[||{\epsilon} - \epsilon_{\SceneScaleDiffusionModel}(t)||_1\Big].
\label{eq:loss_scale}
\end{equation}

At inference time, we sample $\mathbf{S}$ by denoising a noise sample from the standard normal distribution, conditioned on $\mathbf{D}$, and take $s = \text{avg}(\mathbf{S})$ as the isotropic scale factor. The re-scaled depths offer multiple hypotheses for plausible object translation and scale to bring the object from its canonical space to metric camera space.

\subsection{CAD Alignment Diffusion}
\label{sec:alignment}
$\CADAlignmentDiffusionModel$ models a distribution over likely object poses.
We define the 9-DoF transformation as $\mathbf{T} = [\mathbf{R} | \mathbf{t} | \mathbf{s}]$, where $\mathbf{R}\in\text{SO}(3)$ represents rotation, $\mathbf{t}\in\mathbb{R}^3$ translation, and $\mathbf{s} = (s_x, s_y, s_z)^T$ anisotropic scale. 
Instead of modeling this explicit pose representation, which must consider the different spaces for rotation, translation, and scale, we predict the normalized object coordinates (NOCs)~\cite{wang2019normalized} $\mathbf{N}$ of the object.
The NOC representation provides geometric correspondences between the observed object and its canonical coordinate system, enabling solving for $\mathbf{T}$ as well as more generalized learning across different object shapes.
As NOCs $\mathbf{N}$ are structured geometrically similarly to the back-projected points $\mathbf{P}$ of $\mathbf{D}_p$, we condition their prediction on $\mathbf{P}$.

Features $\mathbf{P}_f$ are extracted from condition $\mathbf{P}$ using the 3DGC backbone~\cite{lin2020convolution}.
$\CADAlignmentDiffusionModel$ then operates on a concatenation of the target NOCs with their corresponding per-point features from $\mathbf{P}_f$. 
The objective is to minimize the difference between predicted and ground truth noise, formulated as:

\begin{equation}
\mathcal{L}_{n} = \mathbb{E}_{\epsilon \sim \mathcal{N}(0,I), t} \Big[||\epsilon - \epsilon_{\CADAlignmentDiffusionModel}(t)||_1\Big].
\label{eq:loss_nocs}
\end{equation}

During inference, NOC candidates are sampled from Gaussian noise, conditioned on $\mathbf{P}_f$. 
We can then solve for the 9-DoF object transformation $\mathbf{T}$ from the NOC estimates using a pose solver following a similar approach as presented in CaTGrasp~\cite{wen2021catgrasp}, which is computed via RANSAC~\cite{fischler1981random} to find the transformation from the predicted NOCs and the observed point cloud $\mathbf{P}$.

\subsection{CAD Retrieval Diffusion}
\label{sec:retrieval}
Given our estimated NOCs capturing the visible geometry of the object mapped to its canonical space, we use this information to estimate the 3D shape for retrieval from a CAD database.
As explicit 3D shape representations (e.g., voxels, points) are quite high-dimensional, we employ a latent space representation of objects, with latent codes $\mathbf{z} \in \mathbb{R}^d$.

To compress the shape into latent space, we opt for an architecture similar to ConvONet~\cite{peng2020convolutional}, denoted as $\CADEncodingModel$. We add an additional MLP layer in the encoder to fuse the triplane features at the bottleneck into a global vector with a dimension of $\mathbb{R}^{d}$, and also feed it into the decoder along with interpolated triplane features. We pre-train $\CADEncodingModel$ to encode the CAD models with reconstruction loss following~\cite{peng2020convolutional}. The samples in the latent space in which the $\CADRetrievalDiffusionModel$ learns to sample from are extracted from the encoding of the CAD database using $\CADEncodingModel$.

To encode the information captured by the NOC estimate $\mathbf{N}$, we learn a positional feature embedding that maps point locations of $\mathbf{N}$ from $\mathbb{R}^3$ to $\mathbb{R}^C$ together with a single-layer MLP that serves as the context of the condition for $\CADRetrievalDiffusionModel$.
$\CADRetrievalDiffusionModel$ then learns to denoise the object latent vector at each timestep. We fellow recent works~\cite{ramesh2022hierarchical, chou2023diffusion} and directly optimize for the original denoised latent vector $\mathbf{z}_0$:

\begin{equation}
\mathcal{L}_{z} = ||\CADAlignmentDiffusionModel(\mathbf{z}_t) - \mathbf{z}_0||_1.
\label{eq:loss_latnet}
\end{equation}

During inference, we sample latent vector $\mathbf{z}$ and query the CAD database to retrieve the nearest neighbor based on cosine similarity.

\subsection{Synthetic Dataset Augmentation}
\label{sec:augment}
We train our approach on a synthetic 3D scene dataset 3D-FRONT~\cite{fu20213d} to capture various scene scales, object arrangements, and shapes. 
However, this process can be prone to overfitting because the original 3D-FRONT scene configuration only contains 1,334 unique objects (for our target classes).
To better capture a wider distribution of possible objects and arrangements, we further inject information from other existing large synthetic object databases, such as ShapeNet~\cite{chang2015shapenet}.
We consider the shape database of objects from 3D-FUTURE and those of ShapeNet, yielding 18,229 objects.
We then augment the synthetic 3D scenes by replacing existing furniture with unused CAD models randomly retrieved from the same category in the database. 
This enables learning from more diverse object shapes and arrangements.

Our augmented synthetic dataset comprises $\approx$300k images spanning 6 target categories.

\subsection{Multi-Hypothesis Sampling from an Image}
\label{sec:image_sampling}
Since 3D CAD retrieval and alignment from a single view brings inherent ambiguities in terms of depth-scale, object shape, and pose, our approach addresses the ambiguities by a hierarchical multi-candidate sampling scheme.
Our inference process begins by employing off-the-shelf 2D recognition backbones~\cite{bhat2023zoedepth, xu2023open, eftekhar2021omnidata} to derive depth, mask, and normal estimates from the input image $\mathbf{I}$. Subsequently, we sample in a cascaded fashion from our trained diffusion models, progressing from scene-level reasoning to sampling explicit object representations and ending with the prediction of implicit object representations.

Initially, we sample a set of $n$ potential scales, denoted as $\{s_i\}_{i\in [1,n]}$, from $\SceneScaleDiffusionModel$ based on the predicted depth map $\mathbf{D}$ from $\mathbf{I}$. NOC candidates are then sampled at each scene scale, $\{\mathbf{N}_i\}_{i\in [1,n]}$, given each scaled depth map, from $\CADAlignmentDiffusionModel$. The corresponding 9-DoF transformations  $\{\mathbf{T}_i\}_{i\in [1,n]}$ are computed using a RANSAC-based solver~\cite{fischler1981random}. The possible latent candidates for objects, $\mathbf{z}_i$, are sampled conditionally on $\mathbf{N}_i$ from the $\CADRetrievalDiffusionModel$. We sample $m$ object latent candidates given each $\mathbf{N}_i$ and conduct a nearest neighbor search in the CAD model database to retrieve CAD models based on the cosine similarity between latent vectors.

The hierarchical sampling approach generates $\{n \times m\}$ samples corresponding to various scene scales. For each scene scale $s_i$, a simple hypothesis ranking scheme selects the best candidate by rendering the normals of those CAD models using the solved poses $\mathbf{T}_i$ and computing the similarity between the rendered normal and the machine-estimated normal from RGB input using LPIPS~\cite{zhang2018unreasonable}. The selection criterion is based on the model with the lowest LPIPS error, resulting in $n$ sets of CAD models that align with the probabilistically permuted scenes.

\subsection{Implementation Details}
For $\SceneScaleDiffusionModel$, we use a learning rate of $5e^{-5}$ and batch size 64, on a single RTX a6000 GPU for 2 days.
For $\CADAlignmentDiffusionModel$, we condition on subsampled point clouds with {1024} points, using learning rate $5e^{-5}$ on a single RTX a6000 GPU with a batch size of 96, spanning 3 days per category.
$\CADEncodingModel$ is trained on a single a100 GPU with a batch size of 128 for 3 days, with a learning rate $1e^{-4}$. We then train $\CADRetrievalDiffusionModel$ on an RTX a6000 GPU with batch size 128 and learning rate $1e^{-4}$, for 3 days per category. We detail the condition mechanisms of the diffusion models and the hyperparameters in the Appendix.

%------------------------------------------------------------------------

\section{Experiments}
\label{sec:experiments}

\begin{figure}[!t]
      \centering
      \includegraphics[width=0.95\columnwidth]{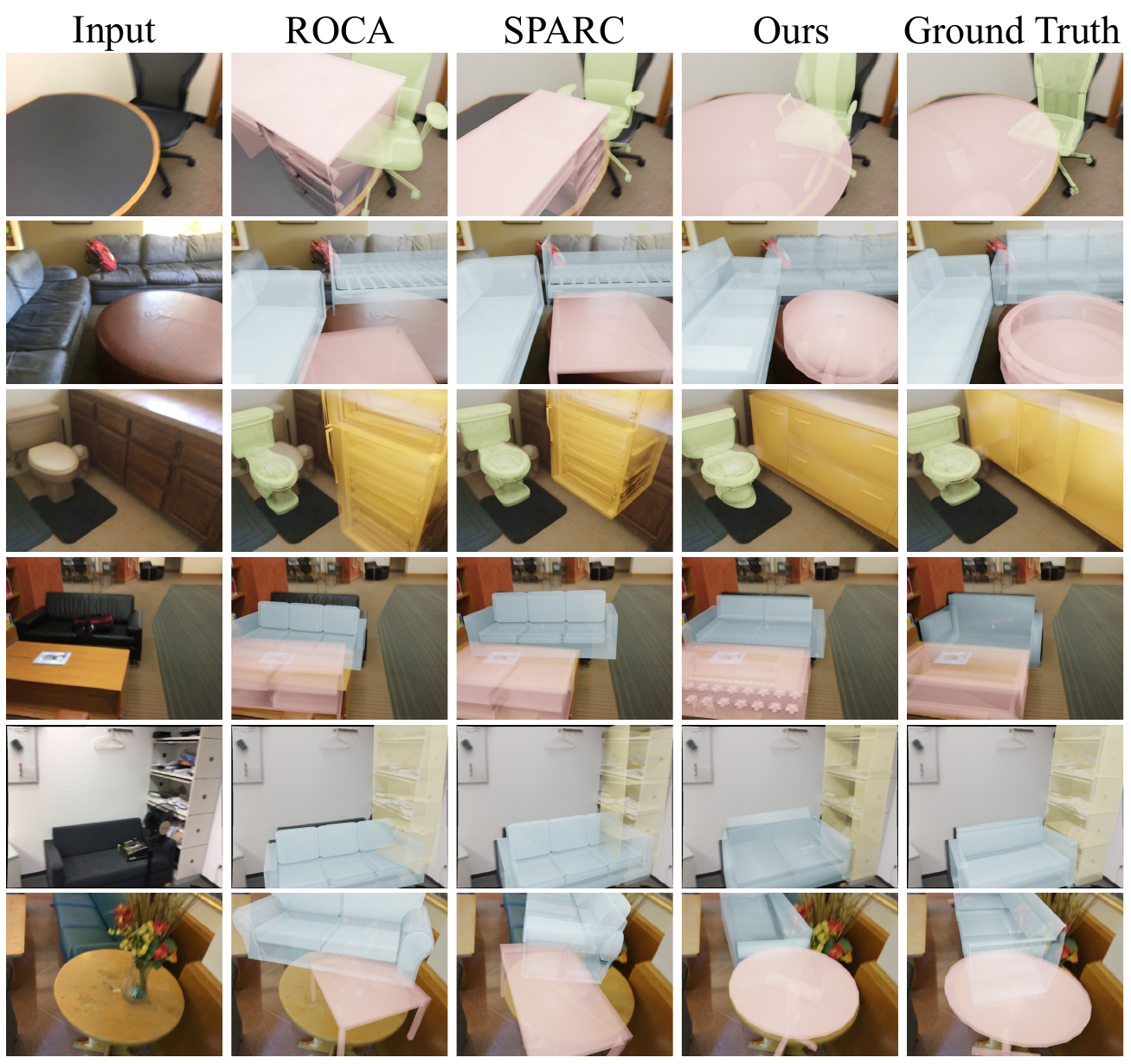}
      \caption{
      \textbf{Qualitative Comparison on ScanNet images~\cite{dai2017scannet,avetisyan2019scan2cad}.} Our weakly-supervised probabilistic approach produces more representative retrieval and alignment, even under strong occlusions (bottom), compared with in-domain supervised methods~\cite{gumeli2022roca,langer2022sparc}.
      }
      \label{figure:qualitative_2}
\end{figure}

We validate our weakly-supervised method on real-world datasets, including ScanNet~\cite{dai2017scannet} and ARKit~\cite{dehghan2021arkitscenes}. As Scan2CAD~\cite{avetisyan2019end} provides ShapeNet~\cite{chang2015shapenet} annotations to ScanNet scenes, we evaluate CAD alignment accuracy and retrieval on the ScanNet25k image data, which contains 5k validation images. Given the challenges in ground truth object shape and pose matching in real data annotations like Scan2CAD, we introduce a probabilistic evaluation protocol for both in-domain supervised approaches and our cross-domain weakly-supervised method. In the absence of CAD annotations for ARKit, we present a qualitative-only evaluation on ARKit in the Fig.~\ref{figure:qualitative_supp_arkit}.

\subsection{Evaluation Metrics}
Our evaluation protocols on ScanNet evaluate the top-n predictions, as is standard practice for evaluating probabilistic methods~\cite{tevet2022human, diller2023cg, guo2022generating, neculai2022probabilistic, chun2021probabilistic} that acknowledges the probabilistic nature of the task and uncovering a distribution that adequately covers the imperfect single-mode annotation.

\subsubsection{Alignment Accuracy}
We introduce a new alignment evaluation protocol for single-view CAD alignment and retrieval.
We observe that previous methods~\cite{gumeli2022roca,langer2022sparc,maninis2022vid2cad} adopted a similar evaluation protocol as Scan2CAD, which operates on reconstructed scenes by aggregating predictions over multiple frames.
Instead, we aim to evaluate the performance given only a single RGB image.
We thus compute alignment accuracy following the same thresholds as in prior work \cite{avetisyan2019scan2cad,gumeli2022roca,langer2022sparc,maninis2022vid2cad}, but only considering a single RGB frame as input: an alignment is correct if the predicted object class is correct, translation error $\leq 20$cm, rotation error $\leq 20^\circ$, and scale ratio $\leq 20\%$. 
As multiple hypotheses should reflect a distribution from which the ground truth is likely to appear, we evaluate multiple hypotheses from each method by evaluating how close the ground truth lies to any hypothesis, that is, the hypothesis with the minimum error as follows:
\begin{equation}
    e = \arccos \Big( \dfrac{\text{tr}\left(\mathbf{\hat{R}}\mathbf{R}^{T}_{gt}\right) - 1}{2} \Big) + ||\mathbf{\hat{t}} - \mathbf{t}_{gt}||_2 + ||\frac{\hat{s}}{{s}_{gt}}-1||_1,
\label{eq:pose_eval}
\end{equation}
where $(\hat{\cdot})$ denotes predictions.

\subsubsection{Retrieval Similarity}
To evaluate the shape retrieval, we establish the retrieval similarity as the L1 Chamfer Distance between point clouds sampled from the retrieved mesh and the corresponding ground truth. We set the candidate CAD models for retrieval as the 1943 models that appear in the Scan2CAD training set. A total of $10k$ points are sampled on the meshes for evaluation. To evaluate our method probabilistically, we report the candidate with the minimum L1 Chamfer Distance according to the ground truth.

\subsubsection{Metric Pose Accuracy}
We calculate average translation (in centimeters), rotation (in degrees), and scaling (in percentage) errors to directly evaluate the metric pose alignment.

\subsubsection{Average L1 Chamfer Distance for Retrieval}
Aside from the standard top-n probabilistic evaluation protocol above, we consider averaging the Chamfer Distance for the retrieval accuracy to illustrate the effectiveness of our diffusion-based retrieval. Note that this metric can penalize for plausible predictions that differ from the single ground truth annotation (e.g., ambiguity in the size of a couch that goes out of the image frame).

\begin{table*}[ht]
    \caption{
        \textbf{Alignment Accuracy on ScanNet} in comparison to state-of-the-art approaches, which require in-domain ground truth supervision. 
        While SPARC~\cite{langer2022sparc} has not been developed for probabilistic reasoning, multiple hypotheses can be generated by sampling different initialization angles (always including those obtained with the officially released initialization angle).
        Evaluation under the same \#hypotheses are colored in \colorbox{tabbestcolor!15}{8}, \colorbox{tabbestcolor!33}{12}, \colorbox{tabbestcolor!55}{16}, and \colorbox{tabbestcolor!88}{20} hypotheses. Our probabilistic approach can outperform fully-supervised methods, while reflecting a more likely distribution of poses.}
    \label{table:main_results}
    \centering
    \resizebox{0.95\textwidth}{!}
        {
        \begin{tabular}{l c c *{6}{>{\centering\arraybackslash}m{0.07\linewidth}} c}
            \toprule% 
            Method                       & \makecell{{in-domain} \\ {supervision}}&\#hypotheses &        bed       &      bkshlf      &      cabinet     &     chair        &      sofa        &       table      & avg$\uparrow$ \\
            \midrule% 
            ROCA~\cite{gumeli2022roca}   & \cmark                             &     {-}     &        11.2      &       11.3       &       12.6       &      35.8        &       8.7        &       9.3        &       14.8    \\
            
            SPARC~\cite{langer2022sparc} & \cmark                             &     {1}     &        26.1      &       21.7       &       27.8       &      47.4        &       25.9       &       19.7       &       28.1    \\
            \rowcolor{tabbestcolor!15}
            SPARC~\cite{langer2022sparc} & \cmark                             &     {8}     &        26.7      &       24.6       &       28.8       &      48.9        &       28.9       &       24.1       &       30.3    \\
            % \rowcolor{\sbestmed} 
            % SPARC~\cite{langer2022sparc} & \cmark                             &     {10}    &        27.3      &       25.1       &       29.1       &      49.0        &       30.4       &       24.5       &       30.9    \\
            \rowcolor{tabbestcolor!33}
            % SPARC~\cite{langer2022sparc} & \cmark                             &     {12}    &        28.6      &       25.6       &       29.8       &      49.3        &       31.2       &       25.0       &       31.6    \\
            \rowcolor{tabbestcolor!55}
            SPARC~\cite{langer2022sparc} & \cmark                             &     {16}    &        29.2      & \underline{26.1} &       30.5       &      49.6        &       31.9       &       26.4       &       32.3    \\
            \rowcolor{tabbestcolor!88}
            SPARC~\cite{langer2022sparc} & \cmark                             &     {20}    &        29.8      & \textbf{27.1}    &       30.8       &      49.8        &       32.7       & \underline{26.8} &       32.8    \\
            % SPARC~\cite{langer2022sparc} & \cmark & {50}  & 31.1  & 27.6 & 31.8 & 48.8 & 31.6 & 26.9 & 32.9 \\
            \midrule
            \textbf{Ours}                & \xmark                             &     {1}     &        13.0      &       4.4        &       9.6        &      22.4        &       13.3       &         4.9      &       11.3    \\
            \rowcolor{tabbestcolor!15}
            \textbf{Ours}                & \xmark                             &     {8}     &        28.6      &       16.7       &       32.8       &      55.0        &       41.1       &         18.6     &       32.1    \\
            % \textbf{Ours}                & \xmark                             &     {10}    &        30.3      &       18.7       &       {34.1}     &      {57.2}      &       {40.7}     &         20.4     &       {33.6}\\
            \rowcolor{tabbestcolor!33}
            \textbf{Ours}                & \xmark                             &     {12}    &        30.4      &       20.2       &       {35.1}     &       {58.5}     &       {41.1}     &         21.6     &       {34.5}  \\
            \rowcolor{tabbestcolor!55}
            \textbf{Ours}                & \xmark                             &     {16}    & \underline{31.1} &       21.7       & \underline{39.4} & \underline{61.3} &\underline{44.9}  &         24.8     &\underline{37.2}\\
            \rowcolor{tabbestcolor!88}
            \textbf{Ours}                & \xmark                             &     {20}    &   \textbf{32.9}  &       24.1       &   \textbf{42.1}  &   \textbf{62.5}  &   \textbf{47.5}  &   \textbf{27.1}  & \textbf{39.4}  \\
            % {\bf Ours} & \xmark & {50} & {39.1} &  {29.1} & {47.4} & {69.3}  &  {54.0} &  {34.0}  & {45.5} \\
            \bottomrule%
        \end{tabular}
        }
\end{table*}

\begin{table*}[hb]
    \caption{
            \textbf{Retrieval Similarity on ScanNet} compared to in-domain supervised state-of-the-arts methods. 
            Note that SPARC~\cite{langer2022sparc} uses the same retrieval provided by ROCA~\cite{gumeli2022roca}. 
            Evaluation under the same \#hypotheses are colored in \colorbox{tabbestcolor!15}{8}, \colorbox{tabbestcolor!33}{12}, \colorbox{tabbestcolor!55}{16}, and \colorbox{tabbestcolor!88}{20} hypotheses. Our approach matches the performance of state of the art with 1 hypothesis, and reflects a likely distribution, retrieving more accurate objects with only a few hypotheses.
            }
    \label{table:main_results_retrieval}
    \centering
    \resizebox{0.95\textwidth}{!}
        {
        \begin{tabular}{l c c *{6}{>{\centering\arraybackslash}m{0.07\linewidth}} c}
            \toprule 
            Method                       & \makecell{in-domain \\ supervision}&\#hypotheses &        bed       &      bkshlf      &      cabinet     &     chair        &      sofa        &       table      & avg$\downarrow$ \\
            \midrule 
            ROCA~\cite{gumeli2022roca}   & \cmark                             & {-}         &       0.087      &      0.089       &      0.130       &      0.100       &      0.099       &       0.132      &   0.106       \\
            SPARC~\cite{langer2022sparc} & \cmark                             & {-}         &       0.087      &      0.089       &      0.130       &      0.100       &      0.099       &       0.132      &   0.106       \\
            \midrule
            \textbf{Ours}                & \xmark                             & {1}         &       0.117      &      0.108       &      0.115       &      0.115       &      0.093       &       0.145      &   0.116      \\
            % {\textbf{Ours}} & \xmark & {5} & 0.081 & 0.076  & 0.092 & 0.086 & 0.076 & 0.103 & 0.086\\
            \rowcolor{tabbestcolor!15}
            \textbf{Ours}                & \xmark                             & {8}         &       0.075      &      0.064       &      0.079       &      0.075       &      0.066       &       0.089      &   0.075      \\
            % \textbf{Ours} & \xmark & {10} & 0.072 & 0.062  & 0.077 & 0.072 & 0.065 & 0.086 & 0.072 \\
            \rowcolor{tabbestcolor!33}
            \textbf{Ours}                & \xmark                             & {12}        &       0.065      &      0.060       &      0.074       &      0.069       &      0.062       &       0.083      &   0.069       \\
            \rowcolor{tabbestcolor!55}
            \textbf{Ours}                & \xmark                             & {16}        &\underline{0.061} &\underline{0.057} &\underline{0.071} &\underline{0.066} &\underline{0.060} &\underline{0.077} &\underline{0.065}\\
            \rowcolor{tabbestcolor!88}
            \textbf{Ours}                & \xmark                             & {20}        & \textbf{0.058}   &   \textbf{0.055} & \textbf{0.069}   & \textbf{0.064}   & \textbf{0.058}   &  \textbf{0.072}  & \textbf{0.063}\\
            \bottomrule
            \end{tabular}}
\end{table*}

\subsection{Comparison to State of the Art}
Tab.~\ref{table:main_results} compares our weakly-supervised method with state-of-the-art single-frame-based methods ROCA~\cite{gumeli2022roca} and SPARC~\cite{langer2022sparc}, both of which require full supervision, on ScanNet~\cite{dai2017scannet, avetisyan2019scan2cad}. 
As the single image setting can contain depth-scale ambiguities, we evaluate multiple hypotheses for methods that allow so.
ROCA is fully deterministic, and so only provides one hypothesis. While SPARC was not developed for multi-hypothesis reasoning, multiple samples can be drawn from different initialization angles (always including the authors' proposed initialization angle).

\OURS{} learns a much more representative distribution, surpassing fully-supervised state of the art with only 8 hypotheses, with notable improvement in both translation and rotation as in Tab.~\ref{table:metric_alignment}, reflecting the quality of our learned distribution. Performance increases with more hypotheses, with a slight saturation of around 20 hypotheses. 
In contrast, the performance of SPARC only improves marginally with increasing the sampled hypotheses.

While our single-hypothesis prediction slightly underperforms compared to fully-supervised approaches, this discrepancy is attributed to the scale-biased nature introduced by the domain gap of synthetic and real-world data. Given inherent imperfectly annotated real-world ground truth, our evaluation considers the likelihood of the ground truth originating from the sampled distribution.

\begin{table}[ht]
    \caption{\textbf{Pose Alignment Metric.} We calculate average translation (in centimeters), rotation (in degrees), and scaling (in percentage) errors to directly evaluate the metric pose alignment. Without any training on real data, our diffusion approach achieves notable improvement in translation and rotation, with the same scale performance.
        }
    \label{table:metric_alignment}
    \begin{center}
    \resizebox{0.46\textwidth}{!}{  
    \begin{tabular}{c c c c c }
    \toprule
       Method                                  & {\#hypotheses}           & Translation$\downarrow$ & Rotation$\downarrow$ & Scale$\downarrow$\\
        \midrule
        ROCA~\cite{gumeli2022roca}             & -                        & 43.5 & 27.9 & 0.19 \\
        SPARC~\cite{langer2022sparc}           & 8                        & 35.1 & 19.5 & 0.19\\
        \textbf{Ours}                          & 8                        & \textbf{27.5} & \textbf{18.6}& 0.19\\
        \bottomrule
    \end{tabular}
    }
    \end{center}
\end{table}

Tab.~\ref{table:main_results_retrieval} evaluates CAD retrieval similarity compared to the state-of-the-art. Since SPARC uses the same deterministic retrieval as ROCA, multiple hypotheses are not available for either method.
With our probabilistic modeling, \OURS{} performs on par with state-of-the-art with one hypothesis and significantly improves with more hypotheses, better reflecting likely shape reconstructions, even with only weak supervision available. 
We further verify the effectiveness of our method by averaging the Chamfer Distance of the retrieved candidates as shown in Tab.~\ref{table:retrieval_avgl1}; even under this metric, our model still outperforms the deterministic baselines.

\begin{table}[hb]
    \caption{\textbf{Average L1 Chamfer Distance for Retrieval.} We average the Chamfer Distance for the retrieval accuracy evaluation. Our approach effectively outperform the deterministic retrieval baselines.
    }
    \label{table:retrieval_avgl1}
    \begin{center}
    \resizebox{0.46\textwidth}{!}{  
    \begin{tabular}{c c c}
    \toprule%
       Method                                  & {\#hypotheses}   & Retrieval Similarity$\downarrow$ \\
        \midrule%
        ROCA~\cite{gumeli2022roca}                & {-}         &      0.106       \\
        SPARC~\cite{langer2022sparc}  & {-}               & 0.106  \\
        \textbf{Ours}                          & 8               & \textbf{0.088} \\
    \bottomrule%
    \end{tabular}
    }
    \end{center}
\end{table}

In Fig.~\ref{figure:qualitative_2}, we show a qualitative comparison of CAD retrieval and alignment on ScanNet images. \OURS{} achieves more accurate object retrieval and alignments across a diverse set of image views and object types due to our effective probabilistic modeling compared with in-domain supervised methods.

\subsubsection{Evaluation on ARKit}
\begin{figure*}
  \centering
  \includegraphics[width=0.93\linewidth]{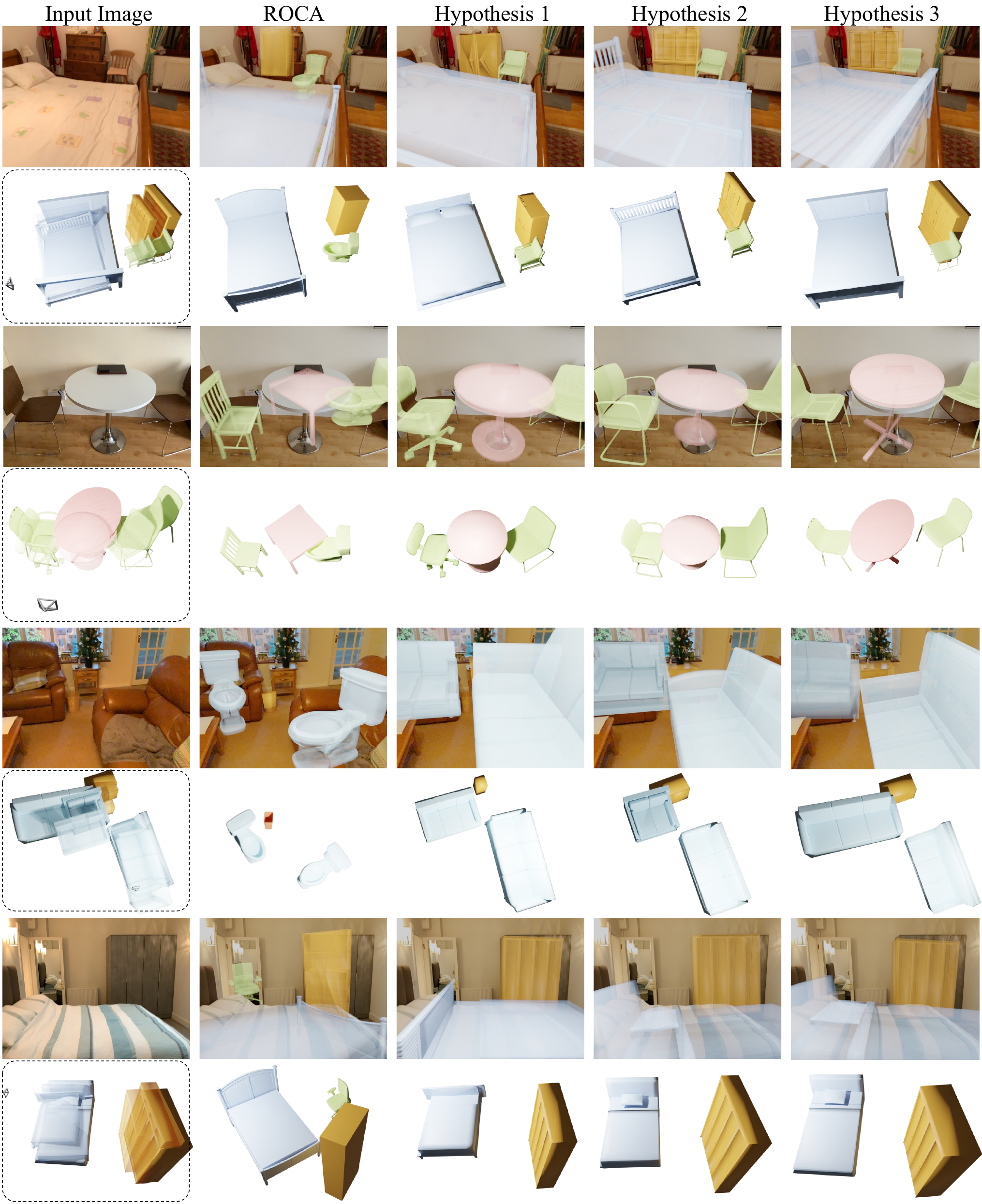}
  \caption{
  \textbf{Qualitative Results on ARKit images.} Our approach presents robust retrieval and alignment to various scenes, reconstructing the scene with multi-feasible sets of object shape and pose pairs given the ambiguities in monocular perception. Dotted: The three hypotheses corresponding to the different scene scales.
  }
  \label{figure:qualitative_supp_arkit}
\end{figure*}

We present qualitative results of our weakly-supervised approach on real-world ARKit~\cite{dehghan2021arkitscenes} data in Fig.~\ref{figure:qualitative_supp_arkit}, comparing against ROCA~\cite{gumeli2022roca}. Our method demonstrates better robustness and accuracy in object retrieval and alignment across diverse images and object types.

\begin{figure*}
  \centering
  \includegraphics[width=0.93\linewidth]{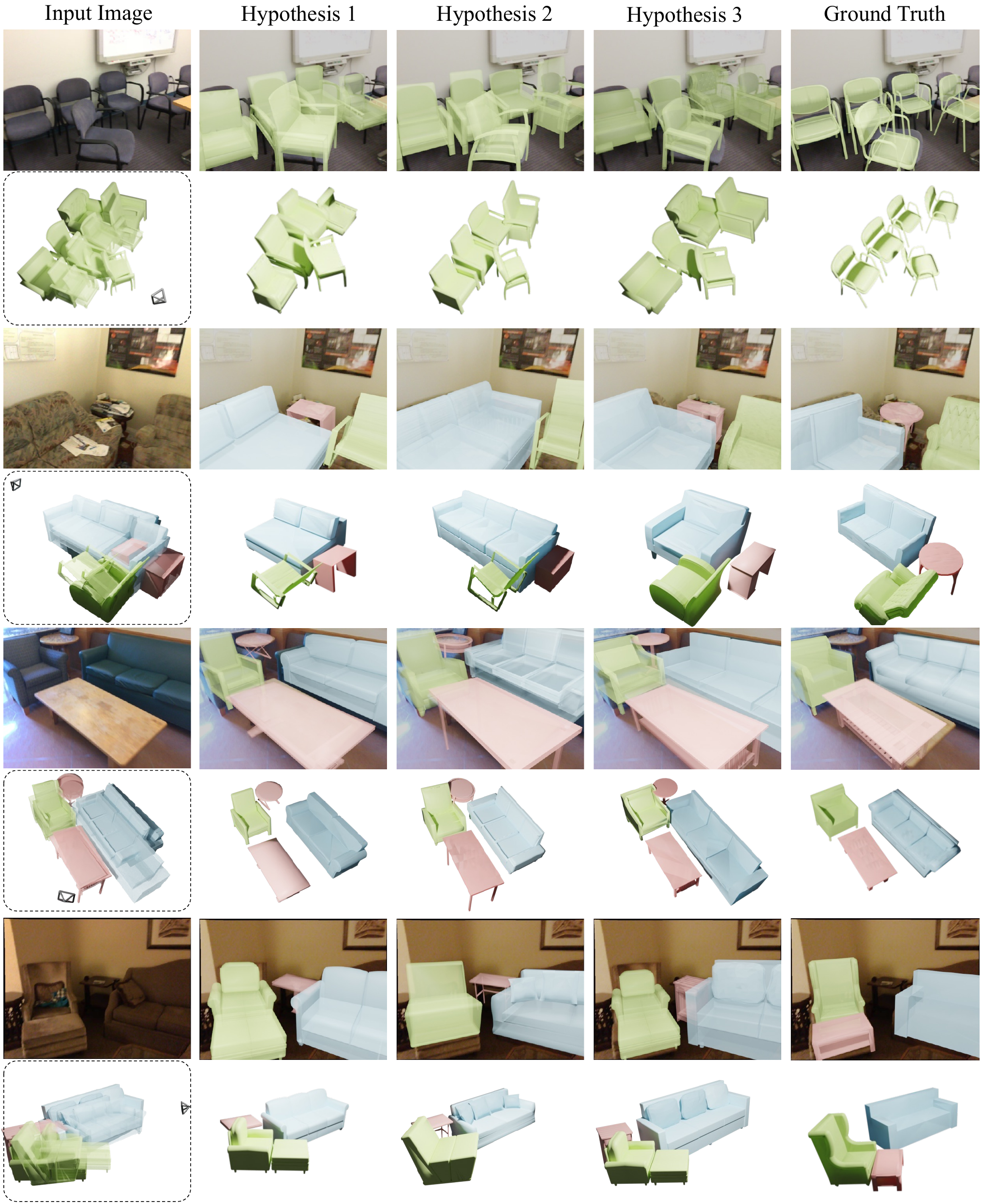}
  \caption{
  \textbf{Qualitative Results on ScanNet images.} Our probabilistic approach shows multi-feasible sets of object shape and pose pairs given the ambiguities in monocular perception. Left-bottom: The two hypotheses corresponding to the smallest and largest scene scale reconstructions follow possible depth-scale ambiguity from the camera view.
  }
  \label{figure:qualitative_mainpaper}
\end{figure*}

\begin{figure}[ht]
      \centering
      \includegraphics[width=0.9\columnwidth]{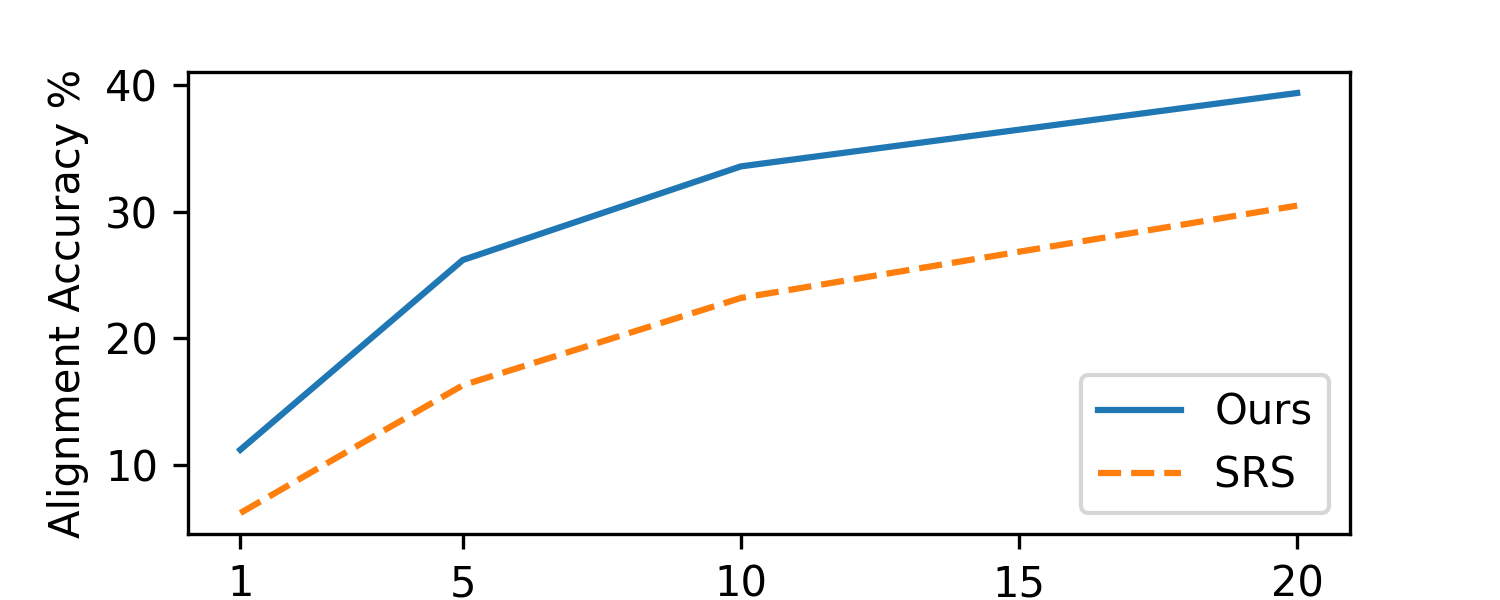}
      \caption{
      \textbf{Scale Sampling Ablation.} Our learned probabilistic model for $\SceneScaleDiffusionModel$ shows much higher sampling efficiency than non-parametric sampling. 
      }
      \label{figure:diffscale_srs_ablation}
\end{figure}

\subsection{Ablations}
\subsubsection{Benefits of Probabilistic Model} 
We consider two alternative deterministic baselines: using the identical UNet architecture as our method but without diffusion, and a PointTransformer-V3~\cite{wu2023point} for NOC prediction. Both baselines are trained on our synthetic train data and evaluated on the real test images. Tab.~\ref{table:nonprob_baselines_short} shows that these deterministic baselines largely suffer from the domain gap and perform inferior to our single-hypothesis model. This indicates the effectiveness and cross-domain robustness of our probabilistic modeling. We detail the class-specific performance of the non-probabilistic baselines in the Appendix.

\begin{table}[ht]
    \caption{
        \textbf{Non-probabilistic Baselines for Alignment Accuracy on ScanNet.} We compare two non-probabilistic baselines that are trained on the synthetic domain, which include a transformer-based backbone~\cite{wu2023point} and a UNet architecture same as Ours. Our method archives much better cross-domain performance on class average alignment metric.
        }
    \label{table:nonprob_baselines_short}
    \centering
    \resizebox{0.47\textwidth}{!}
        {
        \begin{tabular}{l c c c}
            \toprule
            Method                       & \makecell{{in-domain} \\ {supervision}}&\#hypotheses  & avg$\uparrow$ \\
            \midrule
            PointTransformer~\cite{wu2023point} & \xmark    &     {-}     &          6.9    \\
            UNet (same diffusion backbone)  & \xmark           &     {-}     &         10.1    \\
            % \hline
            \textbf{Ours}                & \xmark     &     {1}     &          \textbf{11.3}  \\
            \bottomrule
        \end{tabular}
        }
\end{table}

\subsubsection{Qualitative Evaluation of Probabilistic Retrieval}
\begin{figure}[ht]
      \centering
      \includegraphics[width=0.95\columnwidth]{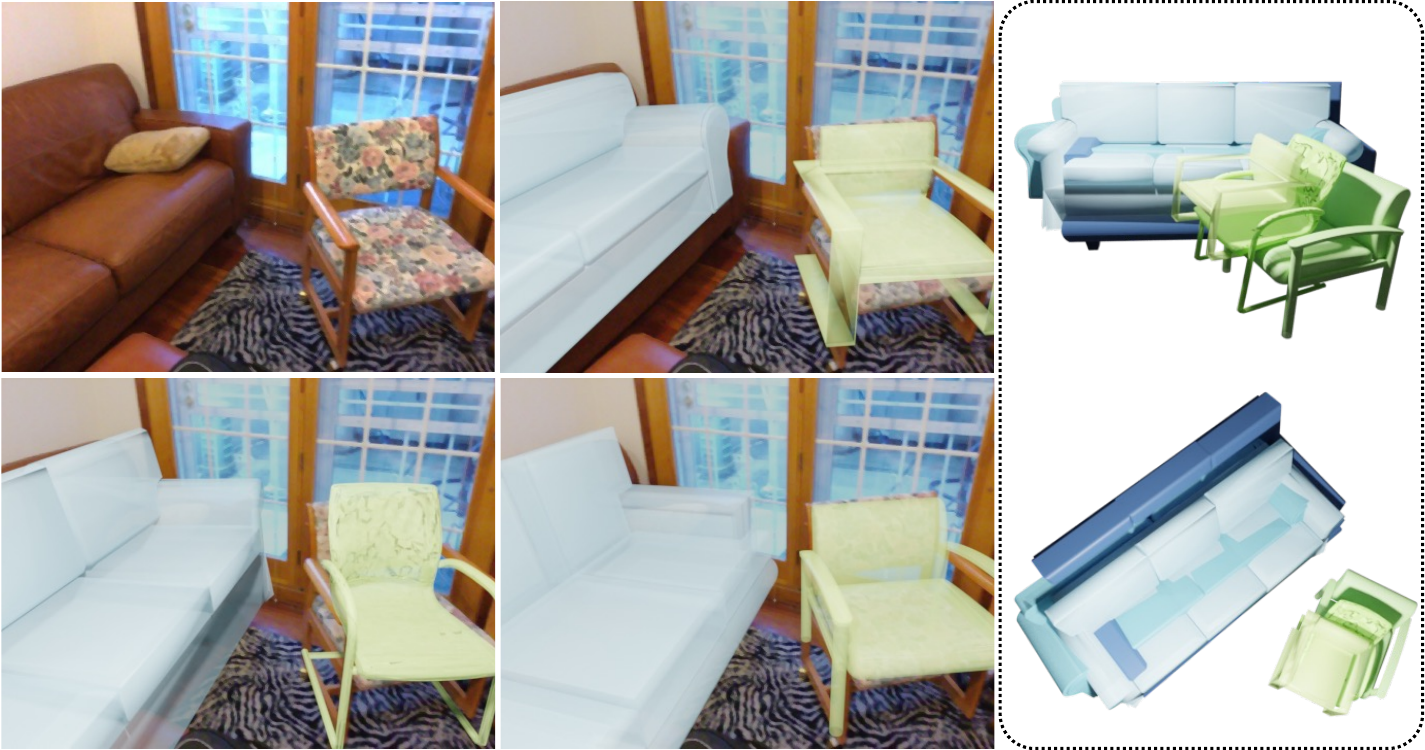}
      \caption{
      \textbf{Multi-hypothesis of Retrieval.} Our approach illustrates various styles of CAD candidates fitted to possible scene scales conditioned on the same input, capturing the solutions to the ambiguities in monocular perception and shape reasoning. Dotted: Hypotheses gathering.
      }
      \label{figure:multihypo_retrieval}
\end{figure}

Fig.~\ref{figure:multihypo_retrieval} demonstrates multiple candidates featuring various styles of chair and sofa that propose various possible scales. In Example 3 of Fig.~\ref{figure:qualitative_supp_arkit}, our probabilistic method successfully identifies potentially different sizes of sofas (single- or double-seater) and occluded cabinets, surpassing deterministic retrieval from ROCA in terms of quality. Example 4 showcases the retrieval of beds with varying lengths, again attributable to occlusion. In Fig.~\ref{figure:qualitative_mainpaper}, examples 2 and 4 further illustrate probabilistic reasoning, showcasing the accommodation of varying shapes and sizes despite limited observations.

\subsubsection{Effectiveness of Our Learned Distributions}
We consider our learned probabilistic distributions in comparison with alternative parametric sampling schemes. We study this for $\SceneScaleDiffusionModel$, as we can employ a straightforward sampling baseline: we estimate the mean $\mu'$ and variance $\sigma'^2$ of scale offsets between predicted and ground truth train depth maps, and instead draw samples from $\mathcal{N}(\mu',\sigma')$ (denoted as `Smart Random Sampling (SRS)' in Fig.~\ref{figure:diffscale_srs_ablation}).
Fig.~\ref{figure:diffscale_srs_ablation} illustrates that our scene scale diffusion model $\SceneScaleDiffusionModel$ outperforms SRS with much higher sampling efficiency.

\subsubsection{Predicting NOCs Instead of Explicit Transformation Parameters $\mathbf{T}$.} 
Tab.~\ref{table:alignment_ablation} shows that our NOC-based proxy for alignment estimation enables much more robust alignment than the direct prediction of the 9-DoF values of $\mathbf{T}$.
The NOC proxy estimation enables more robust estimation through its dense correspondences, which also helps mitigate the potential domain gap between synthetic training data and real-world test scenarios.
We consider two alternatives to explicit $\mathbf{T}$ prediction: `$\CADAlignmentDiffusionModel$ + P' predicts 9-DoF values of $\mathbf{T}$ along with NOCs, and  `$\CADAlignmentDiffusionModel$ + S' predicts only object scale along with NOCs. In `$\CADAlignmentDiffusionModel$ + P', NOCs are solely used for retrieval, whereas in `$\CADAlignmentDiffusionModel$ + S', NOCs are utilized for 6-DoF rotation and translation solving, and retrieval. We evaluate both for the explicit pose prediction as well as for the pose solved by the NOCs via the solver, and select the one with lower error according to Eq.~\ref{eq:pose_eval}.
Our approach to leveraging NOCs and an algorithmic pose solver achieves significantly better cross-domain alignment accuracy. 

\begin{table}[ht]
    \caption{\textbf{Pose Estimation Ablation.} With synthetic data replacement augmentation, our NOC-based predictions from $\CADAlignmentDiffusionModel$ enable more generalized feature learning, resulting in better pose alignment. 
    }
    \label{table:alignment_ablation}
    \begin{center}
    \resizebox{0.44\textwidth}{!}{  
    \begin{tabular}{c  c  c}
    \toprule%
       Method                                  & {\#hypotheses}           & Alignment Accuracy$\uparrow$ \\
        \midrule%
        $\CADAlignmentDiffusionModel$ + P      & 8                        & 16.3 \\
        $\CADAlignmentDiffusionModel$ + S      & 8                        & 19.9 \\
        $\CADAlignmentDiffusionModel$  w/o Aug & 8                        & 25.8 \\
        \textbf{Ours}                          & 8                        & \textbf{32.1} \\
    \bottomrule%
    \end{tabular}
    }
    \end{center}
\end{table}

% 10 hypotheses
% \begin{table}[]
%     \begin{center}
%     \resizebox{0.4\textwidth}{!}{  
%     \begin{tabular}{c | c | c}
%        Method                              & {\#hypotheses}           & Alignment Accuracy$\uparrow$ \\
%         \hline
%         $\CADAlignmentDiffusionModel$ + P  & 10                       & 17.5 \\
%         $\CADAlignmentDiffusionModel$ + S  & 10                       & 21.9 \\
%         \textbf{Ours}                      & 10                       & \textbf{33.6} \\
%     \end{tabular}
%     }
%     \vspace{-3.5mm}
%     \caption{Ablation of predicting explicit pose parameters in $\CADAlignmentDiffusionModel$.
%     \label{table:nocs_parameterization_ablation}
%     \vspace{-0.70cm}
%     \end{center}
% \end{table}

\subsubsection{Effect of Canonical Object Representation for Retrieval}
We consider performing CAD retrieval in the camera space of the image and the canonical space of the CAD database in Tab.~\ref{table:retrieval_ablation}.
We consider a baseline alternative, `$\CADRetrievalDiffusionModel$+PC', which operates on features from the back-projected point cloud in camera space as the retrieval condition.
Learning the shape feature embeddings in canonical space significantly improves retrieval.
\begin{table}[ht]
    \caption{\textbf{CAD Retrieval Ablation.} Learning retrieval in the object canonical space (vs. camera space, denoted as PC) and with synthetic augmentation enables improved retrieval performance. 
    }
    \label{table:retrieval_ablation}
    \begin{center}
    \resizebox{0.4\textwidth}{!}{  
    \begin{tabular}{c c c}
    \toprule
       Method                                  & {\#hypotheses}   & Retrieval Similarity$\downarrow$ \\
        \midrule
        $\CADRetrievalDiffusionModel$ + PC     & 8               & 0.111 \\
        $\CADRetrievalDiffusionModel$ w/o Aug  & 8               & 0.082 \\
        \textbf{Ours}                          & 8               & \textbf{0.075} \\
        \bottomrule
    \end{tabular}
    }
    \end{center}
\end{table}

% 10 hypotheses
% \begin{table}[]
%     \begin{center}
%     \resizebox{0.4\textwidth}{!}{  
%     \begin{tabular}{c | c | c}
%        Method                                  & {\#hypotheses}   & Retrieval Similarity$\downarrow$ \\
%         \hline
%         $\CADRetrievalDiffusionModel$ + PC     & 10               & 0.104 \\
%         $\CADRetrievalDiffusionModel$ w/o Aug  & 10               & 0.081 \\
%         \textbf{Ours}                          & 10               & \textbf{0.072} \\
%     \end{tabular}
%     }
%     \vspace{-3.5mm}
%     \caption{Ablation of learning CAD retrieval in camera space ($\CADRetrievalDiffusionModel$+PC) or learning CAD retrieval without augmenting the synthetic dataset against our design ($\CADRetrievalDiffusionModel$+final).
%     }
%     \label{table:retrieval_ablation}
%     \vspace{-0.70cm}
%     \end{center}
% \end{table}

\subsubsection{Object Replacement Augmentation}
We illustrate the necessity of augmenting the synthetic scene layout with extra shapes (i.e., from ShapeNet~\cite{chang2015shapenet}) in Tabs.~\ref{table:alignment_ablation} and \ref{table:retrieval_ablation}. Both $\CADAlignmentDiffusionModel$ and $\CADRetrievalDiffusionModel$ trained on the augmented synthetic dataset improve significantly compared with `$\CADAlignmentDiffusionModel$ + w/o Aug' and `$\CADRetrievalDiffusionModel$+ w/o Aug' respectively, due to the increased diversity during training.

\subsubsection{Random Neighbors of Deterministic Retrieval}
We conduct experiments retrieving random 8 neighbors of the deterministic retrieval from ROCA which gives 0.093 of L1 Chamfer Distance, which falls behind our diffusion-based retrieval pipeline of 0.075 as shown in Tab.~\ref{table:main_results_retrieval}

\subsection{Limitations}
While leveraging 3D perception through CAD model retrieval and alignment offers a compact representation of the scene, its efficacy for applications requiring exact reconstructions can be hindered by the absence of precise geometric matches in real-world environments. A potential approach to overcome this limitation involves deforming the extracted CAD models to enhance alignment with observations~\cite{ishimtsev2020cad, Di2023UREDU3, uy2020deformation, uy2021joint}. 
Additionally, our approach does not model object relations explicitly, which could limit performance, since indoor scenes are often arranged in coherent global structures. Integrating considerations of scene context alongside object deformation represents a promising direction for achieving more accurate 3D perception.

\section{Conclusion}
\label{sec:conclusion}
We introduce \OURS, the first weakly-supervised probabilistic approach for single-image CAD model retrieval and alignment. By disentangling ambiguities in the monocular perceptual system through individual distribution modeling using diffusion, we effectively address uncertainties in scene depth scale, object shape, and pose. Notably, our diffusion models are trained only on synthetic datasets, and yet outperform the supervised state-of-the-art approach on real-world Scan2CAD image data~\cite{avetisyan2019scan2cad}, achieving a $5.9\%$ improvement with only 8 hypotheses. We envision that this advancement will spur further progress in 3D probabilistic models, as well as 3D perception without necessitating real-world labels.

\begin{acks}
This project is funded by the Bavarian State Ministry of Science and the Arts and coordinated by the Bavarian Research Institute for Digital Transformation (bidt), the ERC Starting Grant SpatialSem (101076253), and the German Research Foundation (DFG) Grant ``Learning How to Interact with Scenes through Part-Based Understanding". We also thank Quan Meng, Lei Li and Hanzhi Chen for the constructive discussions.
\end{acks}

% \clearpage
\bibliographystyle{ACM-Reference-Format}
\bibliography{main}

\appendix
\section{Additional Results}
We present additional ablations and category-specific results for our diffusion-based scene scale, object shape and pose experiments, as well as for the non-probabilistic baselines.

\subsection{Scene Scale Ablation}
Tab.~\ref{table:scale_ablation_fulltable} compares our scene scale diffusion module $\SceneScaleDiffusionModel$ to a baseline approach without the feature extractor $\ScaleFeatureExtractor$. We evaluate the scale prediction accuracy defined as $ \mathbf{e}_s =  \underset{\mathbf{i} \in \{ \mathbf{N}  \} }{\min} || \mathbf{s}_{gt} - \mathbf{s}_i ||_1$. Our design achieves a lower scale error compared to the baseline, indicating improved scale prediction accuracy.

\subsection{CAD Alignment Ablation} 
We present category-specific evaluation results in Tab.~\ref{table:alignment_ablation_fulltable}, corresponding to Tab.3 in the main paper. Tab.~\ref{table:alignment_ablation_fulltable} demonstrates that with synthetic data augmentation and object NOC representation, our CAD alignment diffusion model $\CADAlignmentDiffusionModel$ achieves superior alignment performance across domains for all categories.

\subsection{CAD Retrieval Ablation}
In Tab.~\ref{table:retrieval_ablation_fulltable}, corresponding to Tab.4 in the main paper, we highlight the advantages of learning the retrieval condition in the canonical space, along with the benefits of synthetic data augmentation for the CAD retrieval diffusion model $\CADRetrievalDiffusionModel$.

\subsection{Non-probablistic Baselines for Alignment}
In Tab.~\ref{table:nonprob_baselines}, corresponding to Tab.5 in the main paper, the non-probabilistic baselines suffer from a significant domain gap compared with our approach, verifying the robustness of our diffusion-based weakly supervised strategy.

\subsection{Confidence Measure of Probabilistic Retrieval.}
We estimate model confidence by sample variance, and find that higher-confidence predictions yield improved performance: for object retrieval, performance improves from 0.075 to 0.065 in terms of top-n Chamfer Distance evaluation when considering only high-confidence predictions. We determine high-confidence predictions by setting a 2-$\sigma$ threshold based on the variance. 

\begin{table}[ht]
    \caption{
    \textbf{Scene scale ablation}.
    Encoding the feature from the depth map using $\ScaleFeatureExtractor$ enables better scene scale modeling for the target categories. 
    }
    \centering
    \resizebox{0.47\textwidth}{!}
        {
        \begin{tabular}{l c *{6}{>{\centering\arraybackslash}m{0.08\linewidth}} c}
            \toprule 
       Method                                                     &\#hypotheses     &        bed       &      bkshlf      &      cabinet     &     chair        &      sofa        &       table      & avg$\downarrow$ \\
        \midrule
        $\SceneScaleDiffusionModel$ w/o $\ScaleFeatureExtractor$  & 8               &       0.172      &       0.202      &      0.369       &      0.284       &       0.297      &        0.341     & 0.278 \\
        \textbf{Ours}                                             & 8               &  \textbf{0.133}  & \textbf{0.195}   & \textbf{0.331}   & \textbf{0.251}   & \textbf{0.274}   &\textbf{0.312}    &  \textbf{0.249}      \\
        \bottomrule
       
    \end{tabular}}
    
\label{table:scale_ablation_fulltable}
\end{table}

\begin{table}[ht]
    \caption{
        \textbf{Pose estimation ablation}. With synthetic data replacement augmentation, our NOC-based representation from $\CADAlignmentDiffusionModel$ achieves more generalized feature learning, resulting in better pose alignment for all target categories.
        }
        \label{table:alignment_ablation_fulltable}
    \centering
    \resizebox{0.47\textwidth}{!}
        {
        \begin{tabular}{l c *{6}{>{\centering\arraybackslash}m{0.08\linewidth}} c}
            \toprule% 
            Method                                 &\#hypotheses              &        bed       &      bkshlf      &      cabinet     &     chair        &      sofa        &       table      & avg$\uparrow$ \\
            \midrule%
            $\CADAlignmentDiffusionModel$ + P      & 8                        &       20.5       &       9.4        &        12.3      &      22.9        &      22.1        &         10.5     & 16.3 \\
            $\CADAlignmentDiffusionModel$ + S      & 8                        &       20.5       &       14.3       &        12.9      &      29.3        &      27.8        &         14.6     & 19.9 \\
            $\CADAlignmentDiffusionModel$  w/o Aug & 8                        &       23.0       &       12.8       &        24.5      &      45.3        &      37.3        &         12.0     & 25.8 \\
            \textbf{Ours}                          & 8                        &   \textbf{28.6}  &   \textbf{16.7}  &   \textbf{32.8}  & \textbf{55.0}    &   \textbf{41.1}  &    \textbf{18.6} &  \textbf{32.1} \\
            \bottomrule%
            
        \end{tabular}
        }
\end{table}

\begin{table}[ht]
    \caption{
        \textbf{CAD retrieval ablation}.
        Learning retrieval in the object canonical space (vs. camera space, denoted as `PC') and with synthetic augmentation enables improved retrieval performance for our classes of interest. 
        }
    \label{table:retrieval_ablation_fulltable}
    \centering
    \resizebox{0.47\textwidth}{!}
        {
        \begin{tabular}{l c *{6}{>{\centering\arraybackslash}m{0.08\linewidth}} c}
            \toprule 
       Method                                  &\#hypotheses     &        bed       &      bkshlf      &      cabinet     &     chair        &      sofa        &       table      & avg$\downarrow$ \\
        \midrule
        $\CADRetrievalDiffusionModel$ + PC     & 8               &       0.105      &       0.088      &      0.155       &      0.115       &       0.093      &        0.108     & 0.111 \\
        $\CADRetrievalDiffusionModel$ w/o Aug  & 8               &       0.080      &       0.077      &      0.081       &      0.085       &       0.069      &        0.097     & 0.082 \\
        \textbf{Ours}                          & 8               &  \textbf{0.075}  & \textbf{0.064}   & \textbf{0.079}   & \textbf{0.075}   & \textbf{0.066}   &\textbf{0.089}    &  \textbf{0.075}      \\
        \bottomrule
        
    \end{tabular}}
\end{table}
\begin{table}[ht]
    \caption{
        \textbf{Non-probabilistic Baselines for Alignment Accuracy on ScanNet.} We compare two non-probabilistic baselines that are trained on the synthetic domain, which include a transformer-based backbone~\cite{wu2023point} and a UNet architecture same as Ours. Our method archives much better cross-domain performance on class average alignment metric.
        }
    \label{table:nonprob_baselines}
    \centering
    \resizebox{0.47\textwidth}{!}
        {
        \begin{tabular}{l c c *{6}{>{\centering\arraybackslash}m{0.07\linewidth}} c}
            \toprule% 
            Method                       & \makecell{{in-domain} \\ {supervision}}&\#hypotheses &        bed       &      bkshlf      &      cabinet     &     chair        &      sofa        &       table      & avg$\uparrow$ \\
            \midrule% 
            PointTransformer~\cite{wu2023point} & \xmark                             &     {-}     &        5.38      &       3.14       &       1.66       &      19.89        &       10.44       &       1.09       &       6.93    \\
            UNet (same diffusion backbone)  & \xmark                             &     {-}     &        12.72      &       \textbf{6.99}       &       1.42       &      19.54        &       \textbf{19.68}        &       0        &       10.06    \\
            \textbf{Ours}                & \xmark                             &     {1}     &        \textbf{13.0}      &       4.4        &       \textbf{9.6}        &      \textbf{22.4 }       &       13.3       &         \textbf{4.9}      &       \textbf{11.3} \\
            \bottomrule%
        \end{tabular}
        }
\end{table}

\section{Data Preparation}
\subsection{Object Mesh Pre-processing}
To encode CAD models into a compact latent space using $\CADEncodingModel$, we first canonicalize the original meshes from 3D-FUTURE~\cite{fu20213d} and ShapeNet~\cite{chang2015shapenet} to have a normalized scale and consistent orientation. We save the scaling factor between the original and the canonicalized object for the synthetic data augmentation. Subsequently, we transform them into watertight meshes following the mesh-fusion process proposed in~\cite{stutz2020learning}. 

\subsection{Synthetic Dataset Augmentation}
We leverage the shape databases of objects from both 3D-FUTURE and ShapeNet, with a total of 18,229 objects. We substitute the original furniture of our target class within the 3D-FRONT layout. To achieve this, we randomly select an unused CAD model from the corresponding category in the database. Then, we scale the chosen object using the scaling factor of the original object, ensuring the preservation of object size balance in the augmented scenes. This augmentation process facilitates the learning from a more diverse set of object shapes and arrangements.

\subsection{Synthetic Data Rendering}
We employ BlenderProc~\cite{Denninger2023} for our synthetic data rendering, generating RGB images, depth maps, and masks. Camera views are sampled by considering ray intersections with a minimum coverage of $15\%$ of the object of interest, highlighting our focus on object-centric learning. Additionally, an off-the-shelf depth estimator, such as Zoedepth~\cite{bhat2023zoedepth}, is used to obtain depth estimates over the rendered color images. This information is necessary for calculating the target scale $s$ by comparing predicted $\mathbf{D}$ and rendered depth values $\mathbf{D}_\textrm{gt}$ regarding Eq.5.

\begin{figure}
  \centering
  \includegraphics[width=0.95\linewidth]{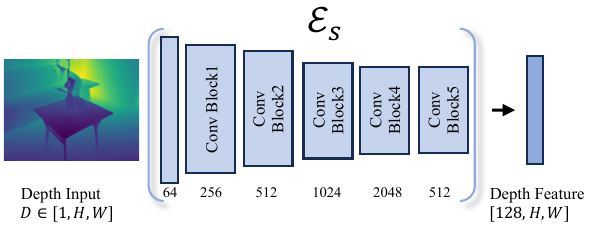}
  \caption{
  \textbf{Overview of the Depth Feature Extractor $\ScaleFeatureExtractor$.} The feature extractor takes as input the predicted depth map, and processes it with hierarchical convolution blocks following~\cite{long2015fully}, and outputs the extracted feature with the same spatial dimension as input. We indicate the feature dimension under each block.
  }
  \label{figure:scale_featextractor}
\end{figure}
\begin{figure}
  \centering
  \includegraphics[width=0.95\linewidth]{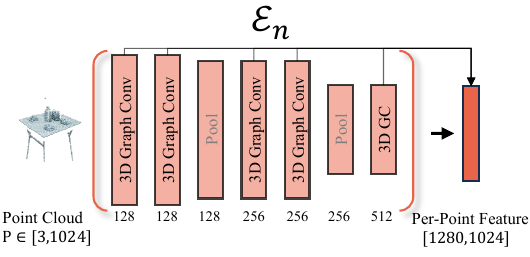}
  \caption{
  \textbf{Overview of the Object Point Cloud Feature Extractor $\DepthFeatureExtractor$.} The feature extractor takes the back-projected object point cloud as input, and extracts the per-point feature using GCN3D~\cite{lin2020convolution}. We indicate the per-point feature dimension below each 3D graph convolution block.
  }
  \label{figure:pc_featextractor}
\end{figure}

\section{Further Implementation Details}

\subsection{Network Architecture}

\subsubsection{Depth feature extractor $\ScaleFeatureExtractor$}
Fig.~\ref{figure:scale_featextractor} diagrams the depth feature extractor network. We adopt the FCN-ResNet50~\cite{long2015fully} and adjust its first convolution layer to take a single channel depth map as input, and use the feature before the final classification layer as the condition for the scene scale diffusion model $\SceneScaleDiffusionModel$.

\subsubsection{Object point cloud feature extractor $\DepthFeatureExtractor$}
We utilize the GCN3D~\cite{lin2020convolution} as feature extraction backbone given the object point cloud $\mathbf{P}$ back-projected from masked depth map $\mathbf{D}_p$. Fig.~\ref{figure:pc_featextractor} shows the object per-point feature extraction process, the feature then serves as input to the CAD alignment diffusion model $\CADAlignmentDiffusionModel$.

\subsubsection{NOC point embedding $\NOCSFeatureEmbedding$}
Given that NOCs inherently capture the observed object in its canonical space, we focus on learning the features of NOCs capable of distinguishing intra-class shape variations. To achieve this objective, we directly learn the feature embedding of NOCs through trigonometric mappings. Subsequently, a single-layer MLP is employed to project the per-point features into a $512$-dimensional space. This NOC feature then serves as a condition for the CAD retrieval diffusion model $\CADRetrievalDiffusionModel$.

\subsubsection{Condition Mechanism} For $\SceneScaleDiffusionModel$ and $\CADAlignmentDiffusionModel$, we concatenate the condition feature with the noised target vector as input to the diffusion UNet. For $\CADRetrievalDiffusionModel$, we learn the positional feature embedding given the condition and use cross-attention to inject the feature at every diffusion UNet block following~\cite{cheng2023sdfusion} which maps explicit shape geometry with object code in the latent space.

\subsubsection{Network Architecture Choice} Our design can also be extended to diffusion-based autoregressive backbone such as DiT~\cite{peebles2023scalable}. An essential modification involves properly tokenizing the target. By initially encoding the target vector and adopting the patchify operation employed in DiT, our approach aligns with the principles of the aforementioned architecture.

\subsection{Test Split Generation}
We introduce a new evaluation protocol for single-view CAD model retrieval and alignment on the ScanNet~\cite{dai2017scannet, avetisyan2019scan2cad}. Our per-frame test split is designed based on the validation set of the ScanNet25k image dataset, aligning with previous methods~\cite{gumeli2022roca, langer2022sparc}. For each target category, we first filter out the frames in which the centroid of the target object falls beyond the image plane. To avoid penalizing the heavily occluded objects, frames where the visible mask of the target object occupies less than $10\%$ of the total image plane are also excluded. The visible masks are obtained by comparing the rendered depth map using ground truth pose with the original sensory depth map. In total, our test split contains around 2.5k images across the 6 target categories.

\subsection{Hyperparameters for Pose Solver}
For RANSAC pose, we use OpenCV estimateAffine3D with confidence threshold 0.999 and RANSAC threshold 0.005. We randomly select 2000 4-point pairs to solve for alignment, using the transformation with the largest inlier ratio~\cite{wen2021catgrasp}.

\subsection{Runtime Analysis}
Our method takes ~12.09s for an image on a single RTX 3080 GPU, with 0.64s for scene scale sampling, 7.63s for NOC sampline (58ms for RANSAC alignment solver), and 3.82s for object latent sampling (1.52 ms for nearest neighbor retrieval).

\section{Used Open-Source Libraries} 
Our data pre-processing, model training and inference paradigm leverage several open-source libraries. Blenderproc~\cite{Denninger2023} is employed for rendering synthetic data. The implementation of our model uses PyTorch. For the multi-hypothesis ranking scheme during inference, which includes rendering the retrieved CAD models using predicted poses, we utilize a PyTorch3D renderer~\cite{ravi2020accelerating}.

\clearpage

\end{document}